\documentclass{article}

\usepackage{arxiv}

\usepackage[utf8]{inputenc} 
\usepackage[T1]{fontenc}    
\usepackage{hyperref}       
\usepackage{url}            
\usepackage{booktabs}       
\usepackage{amsfonts}       
\usepackage{nicefrac}       
\usepackage{microtype}      
\usepackage{lipsum}
\usepackage{graphicx}
\usepackage{changes}
\usepackage{verbatim}
\usepackage{algorithm}
\usepackage{amsmath}
\usepackage{algorithmic}
\usepackage{amsfonts,amssymb}
\graphicspath{ {./images/} }

\title{Efficient Apple Maturity and Damage Assessment: A Lightweight Detection Model with GAN and Attention Mechanism}

\author{
 Yufei Liu \\
  College of Information and Electrical Engineering\\
  China Agricultural University\\
  Beijing,China 100083\\
   \And
 Manzhou Li \\
  College of Plant Protection\\
  China Agricultural University\\
  Beijing,China 100083\\
  \And
 Qin Ma \\
  College of Information and Electrical Engineering\\
  China Agricultural University\\
  Beijing,China 100083\\
}

\begin{document}
\maketitle
\begin{abstract}
This study proposes a method based on lightweight convolutional neural networks (CNN) and generative adversarial networks (GAN) for apple ripeness and damage level detection tasks. Initially, a lightweight CNN model is designed by optimizing the model's depth and width, as well as employing advanced model compression techniques, successfully reducing the model's parameter and computational requirements, thus enhancing real-time performance in practical applications. Simultaneously, attention mechanisms are introduced, dynamically adjusting the importance of different feature layers to improve the performance in object detection tasks. To address the issues of sample imbalance and insufficient sample size, GANs are used to generate realistic apple images, expanding the training dataset and enhancing the model's recognition capability when faced with apples of varying ripeness and damage levels. Furthermore, by applying the object detection network for damage location annotation on damaged apples, the accuracy of damage level detection is improved, providing a more precise basis for decision-making. Experimental results show that in apple ripeness grading detection, the proposed model achieves 95.6\%, 93.8\%, 95.0\%, and 56.5 in precision, recall, accuracy, and FPS, respectively. In apple damage level detection, the proposed model reaches 95.3\%, 93.7\%, and 94.5\% in precision, recall, and mAP, respectively. In both tasks, the proposed method outperforms other mainstream models, demonstrating the excellent performance and high practical value of the proposed method in apple ripeness and damage level detection tasks. This study provides an effective intelligent solution, which might applicable for the agricultural industry, and future research will continue to optimize the model structure and training strategies, exploring more application scenarios to meet the demands of industry development.
\end{abstract}


\section{Introduction}
\label{sec: introduction}

In recent years, the apple industry, as an important part of agriculture, has experienced rapid development worldwide. However, pests and diseases can cause reduced apple productivity, leading to significant economic losses\cite{Khan2022DeepDA}. To ensure the quality and safety of apples, it is crucial to accurately detect and classify their maturity and damage levels. The maturity and damage levels of apples directly affect their market value and consumer willingness to purchase, so accurate classification and detection of apples, as well as determining their maturity and defects, have substantial economic value and practical significance. Traditional methods for detecting apple maturity and damage levels primarily rely on manual vision and experience, resulting in low efficiency, limited accuracy, and susceptibility to human factors. Therefore, how to leverage modern technology to improve the accuracy and efficiency of apple maturity and damage detection has become a hot research topic.

In recent years, deep learning technology has achieved significant results in the field of computer vision, particularly convolutional neural networks (CNNs), which have demonstrated excellent performance in image recognition and object detection tasks\cite{brain, skin, ct, lin}. Existing research shows that using deep learning methods for agricultural product classification and detection is an effective approach \cite{pear, maize, leaf, wheat, seedling}. For instance, Zhu et al. proposed an Apple-Net based on YOLOv5 to detect apple leaf diseases. The experimental results show that the mAP@0.5 value of their proposed model is 95.9\%, surpassing SSD, Faster RCNN, YOLOv4, and the original YOLOv5. Furthermore, Apple-Net achieved an accuracy of 93.1\%. Therefore, Apple-Net can effectively detect apple leaf diseases; however, their model did not consider model parameters or lightweight design, so training the model takes a long time and cannot meet the needs of real-time detection in farms\cite{zhu2023apple}. In other words, deep learning models often require substantial computational resources and storage space, posing a difficult challenge for real-time applications like apple classification and detection. Therefore, researching a lightweight deep learning model to meet real-time requirements is of significant research value. In existing studies, many deep learning-based classification and detection methods for crops and plants have been developed. Saleem, Muhammad Hammad et al. conducted a comparative evaluation of deep learning-based plant disease classification, comparing various CNN networks based on performance indicators such as validation accuracy/loss, F1 scores, and the number of required epochs. They found that Xception, improved GoogLeNet, and cascaded versions of AlexNet and GoogLeNet models achieved the highest validation accuracy and F1 scores in their respective categories, providing convenience for future research. Although the Xception model provided the best results, it required a large amount of time to complete each epoch\cite{sal2020plant}. Phan, Quoc-Hung et al. used YOLOv5 and ResNet to detect the maturity and damage levels of tomatoes, obtaining overall accuracies of 98\%, 98\%, 97\%, and 97\% for ResNet-50, EfficientNet-B0, Yolov5m, and ResNet-101 models, respectively. However, their combination of Yolov5m and Yolov5m with ResNet50, ResNet-101, and EfficientNet-B0 resulted in more model parameters\cite{sal2020plant}. Abbas, Irfan et al. used four convolutional neural network (SqueezeNet, EfficientNet-B3, VGG-16, and AlexNet) CNN models to identify leaf scorch disease in strawberry plants, with the EfficientNet-B3 model achieving the highest accuracy\cite{Abb2021stra}. Dadashzadeh, Mojtaba et al. developed a stereo vision system for distinguishing rice plants from weeds and further differentiated two types of weeds in rice fields using artificial neural networks (ANN). They ultimately achieved an accuracy of 88\%; however, the average processing time per frame reached 0.6 seconds, which could not adequately meet real-time requirements\cite{Abb2021stra}.

Li et al. proposed a DeepLabV3+ semantic segmentation network model based on Atrous Spatial Pyramid Pooling (ASPP) modules for apple leaf disease detection and developed a related application. However, after testing, it was found that processing each disease image took approximately 44 seconds\cite{Li2023dia}. Nasiri, Amin et al. proposed a Convolutional Neural Network (CNN) framework based on the VGG architecture to automatically identify grape varieties using leaf images, ultimately achieving 99\% accuracy\cite{Na2021au}. Chen et al. proposed a lightweight model based on the YOLOv7 network model for citrus detection, introducing a small object detection layer, lightweight convolution, and CBAM (Convolutional Block Attention Module) attention mechanism, ultimately achieving a mean average precision (mAP@0.5) of 97.29\%. However, higher accuracy came at the expense of longer inference times\cite{chen2022a}.

Qi et al. proposed an MC-LCNN for medicinal chrysanthemum detection, which used MC-ResNetv1 and MC-ResNetv2 as the main networks, ultimately achieving 93.06\% accuracy. However, practical environment considerations such as overlap and occlusion pose challenges for model robustness\cite{qi2022me}. Yao et al. established the first dataset on kiwifruit and used YOLOx to detect kiwifruit leaf diseases, ultimately achieving 95\% accuracy\cite{yao2022two}. Fraiwan, Mohammad et al. used deep learning algorithms to distinguish four states of corn—disease one, disease two, disease three, and healthy—achieving 98.6\% accuracy. However, the dataset used in the experiment was limited, and the actual application of corn diseases is not limited to three types, so the model could not meet the actual needs of farmers\cite{yao2022two}.

In response, Craze, Hamish A. et al. used an augmented dataset containing 18,656 images of mixed corn diseases to accurately identify gray leaf spot (GLS) disease in corn, with the final GLS classification experiment achieving 94.1\% accuracy. However, when tested in real scenarios, the accuracy dropped significantly to only 55\%\cite{cra2022deep}. Li et al. proposed a new neural network, YOLO-JD, for identifying ramie diseases based on YOLO. It integrated three new modules—Sand Clock Feature Extraction Module (SCFEM), Deep Sand Clock Feature Extraction Module (DSCFEM), and Spatial Pyramid Pooling Module (SPPM)—ultimately surpassing other models with a 96.63\% mAP\cite{li2022yolo}. To reduce losses caused by rice diseases, Latif, Ghazanfar et al. proposed a deep convolutional neural network (DCNN) transfer learning method based on VGG, achieving a 0.96 accuracy after data augmentation\cite{la2022deep}. Kanda, Paul Shekonya et al. used ResNet18, ResNet34, ResNet50, ResNet102, and ResNet152 to identify tomato leaf diseases, with ResNet152 performing the best at 99\% accuracy\cite{ka2022to}.

Cheng et al. proposed a high-performance wheat disease detection method based on location information, with results showing that the proposed location attention block could improve ResNet's accuracy by up to 2.7\%. However, this led to an increased number of model parameters and longer inference times\cite{Xiao2021de}. Wang et al. designed a pest and disease intelligent detection platform based onYOLOx\_s to assist in wine production, achieving an overall accuracy of 92\%, significantly reducing manual labor costs\cite{wang2023de}. Li et al. proposed an ensemble model based on YOLO and Faster-RCNN to detect 37 types of pests and diseases, ultimately achieving an mAP of 85\%, with detection accuracy still requiring improvement\cite{wang2023de}.

Yebasse, Milkisa et al. used visualization methods to classify coffee diseases, employing three visualization techniques: Grad-CAM, Grad-CAM++, and Score-CAM, ultimately achieving 98\% accuracy\cite{ye2021coff}. Yee-Rendon, Arturo et al. proposed a method for identifying tobacco mosaic virus (TMV) and pepper yellow vein virus (PHYVV) on Mexican pepper leaves using image processing and deep learning classification models. Additionally, data augmentation was employed to prevent model overfitting, ultimately achieving 98\% accuracy. However, the dataset used was relatively small, limiting the model's generalizability\cite{yee2021ana}. Xiao et al. proposed a CNN algorithm based on ResNet50 for strawberry disease identification, comparing it to VGG-50 and GoogLeNet models. Results showed that the ResNet50-based algorithm performed the best, achieving a training dataset accuracy of 98\%. However, their dataset was small and did not include healthy strawberries, making the resulting model insufficient for practical needs\cite{Xiao2021de}.

While these methods achieved some success in terms of accuracy and efficiency, they still have issues and shortcomings, such as a large number of model parameters, high computational requirements, and poor real-time performance. To address these problems, this paper proposes an apple ripeness and damage detection method based on lightweight convolutional neural networks and generative adversarial networks (GANs). The main contributions and innovations of this paper are as follows:

A lightweight convolutional neural network (CNN) model is proposed, effectively reducing the number of model parameters and computational requirements while enhancing the model's real-time performance in practical applications. By optimizing the depth and width of the model and adopting advanced model compression techniques, the lightweight design of the model is achieved, reducing computational resource demands while ensuring classification and detection accuracy.

Generative adversarial networks (GANs) are employed to augment the dataset, improving the model's generalization. To address the issues of sample imbalance and insufficient sample quantity in apple ripeness and damage detection tasks, realistic apple images are generated using GANs, thereby expanding the training dataset and enhancing the model's recognition capabilities for apples with varying ripeness and damage levels.

The object detection network is applied to annotate damaged locations on damaged apples, improving the accuracy of damage degree detection. By combining the object detection network with the lightweight CNN, precise annotation of damaged apples is achieved, contributing to a more accurate assessment of apple damage degrees and providing a more precise basis for decision-making.

The proposed method achieves favorable results in apple ripeness and damage degree detection tasks, demonstrating its effectiveness and practicality. Experiments on real datasets validate the advantages of the proposed lightweight CNN and GANs in improving apple classification and detection accuracy while reducing computational resource requirements.

\section{Method}
\label{sec: Method}

\subsection{Attention Mechanism}
Convolutional Neural Networks (CNNs) sometimes encounter insufficient local information capture when dealing with visual tasks, which can impact the performance of the model. To address this issue, a Convolutional Block Attention Module (CBAM) based on attention mechanisms is proposed in this section to enhance the representation learning capability of CNNs. CBAM is a model optimization method based on channel and spatial attention mechanisms that can be used to improve the accuracy and generalization capability of convolutional neural networks.

\subsubsection{Channel Attention}
\begin{figure}[H]
    \includegraphics[width=14cm]{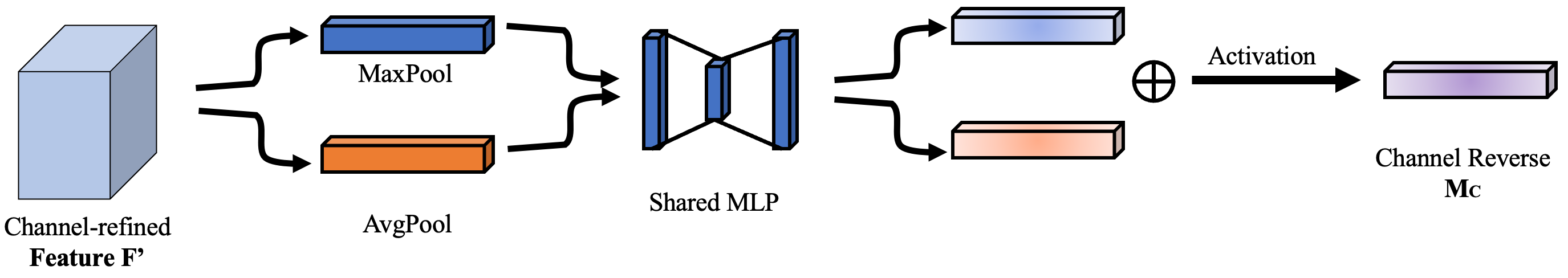}
    \caption{Illustration of channel attention method.}
    \label{fig: ca}
\end{figure}

In the basic block of ResNet, channel attention is primarily used to adjust the weights among different channels, as shown in Figure \ref{fig: ca}.

CBAM calculates the importance of each channel using a global average pooling operation. Let the input feature map be $x$ with $C$ channels; then, the importance score for the $i-th$ channel can be represented as:

\begin{equation}
    f_{channel}(x_{i})=\frac{1}{H \times W}\sum_{j=1}^{H}\sum_{k=1}^{W}x_{ijk}
\end{equation}

where $H$ and $W$ represent the height and width of the input feature map, respectively. The score matrix $f_{channel}(x)$ can be calculated using fully connected layers and activation functions:

\begin{equation}
    M_{channel}(x)=\sigma(W_{2}ReLU(W_{1}f_{channel}(x)))
\end{equation}

where $W_1$ and $W_2$ represent the weight matrices of the fully connected layers, ReLU represents the activation function, and $\sigma$ represents the sigmoid function. The score matrix $M_{channel}(x)$ represents the importance of each channel, which can be obtained by multiplying it with the input feature map to obtain the adjusted feature map:

\begin{equation}
    x_{channel}(x)=M_{channel}(x) \cdot x
\end{equation}

\subsubsection{Spatial Attention}
In the basic block of ResNet, spatial attention is primarily used to adjust the weights among different spatial positions.

CBAM calculates the importance of each spatial position using fully connected layers and convolution operations. Let the input feature map be $x$ with height $H$ and width $W$; then, the importance score for position $(i, j)$ can be represented as:

\begin{equation}
    f_{spatial}(x_{ij})= \sigma (W_{2}ReLU(W_{1}x_{ij}))
\end{equation}

where $W_1$ and $W_2$ represent the weight matrices of the fully connected layers, ReLU represents the activation function, and $\sigma$ represents the sigmoid function. Next, an interpretation and derivation of the above formula will be provided.

First, consider the function of the Squeeze module. Its input is the feature map $X \in \mathbb{R}^{W \times H \times C}$, and its output is the global statistical information of the feature map, i.e., the channel attention vector $z \in \mathbb{R}^{C'}$. This process can be implemented through global average pooling and two fully connected layers:

\begin{equation}
    \begin{split}
    z&=f_{MLP}(AvgPool(X))\\
    &=f_{MLP}(\frac{1}{W\times H}\sum_{i=1}^{W}\sum_{j=1}^{H}X_{i,j})\\
    &=f_{MLP}(u)
    \end{split}
\end{equation}

where $\mathrm{AvgPool}$ is the global average pooling operation, and $u$ is the feature vector after average pooling. $f_{\mathrm{MLP}}$ is a multilayer perceptron composed of two fully connected layers, and its output is the channel attention vector $z$.

Next, consider the function of the Excitation module. Its input is the feature map $X \in \mathbb{R}^{W \times H \times C}$ and the channel attention vector $z \in \mathbb{R}^{C'}$, and its output is the weighted feature map $Y \in \mathbb{R}^{W \times H \times C}$. This process can be represented as:

\begin{equation}
    Y=g_{MLP}(z,X)
\end{equation}

where $g_{\mathrm{MLP}}$ is a multilayer perceptron, and its input is $z$ and $X$, and its output is the weighted feature map $Y$. Specifically, $g_{\mathrm{MLP}}$ consists of two fully connected layers and a sigmoid activation function:

\begin{equation}
    \begin{split}
    g_{MLP}(z,X)&=sigmoid(W_{2}\delta(W_{1}[z,X]))\\
    &=sigmoid(W_{2}\delta(W_{1}[f_{MLP}(AvgPool(X)),X]))\\
    \end{split}
\end{equation}

where $[z,X]$ represents concatenating $z$ and $X$ along the channel dimension, $\delta$ represents the activation function (ReLU is used here), and $W_{1}$ and $W_{2}$ are the weight parameters of the two fully connected layers.

Integrating the two modules mentioned above, the entire process of CBAM can be represented as:

\begin{equation}
    Y=g_{MLP}(f_{MLP}(AvgPool(X)),X)
\end{equation}

In other words, CBAM extracts the global statistical information of the feature map through the Squeeze module and calculates the weighted feature map through the Excitation module, thereby enhancing the representation capability of the feature map. Applying CBAM to classification networks can yield favorable results for several reasons:

\begin{enumerate}
    \item CBAM introduces channel attention mechanisms and spatial attention mechanisms, enabling the model to more accurately focus on important feature information in images.
    \item  Channel attention mechanisms allow the network to pay more attention to the importance between channels, removing redundant information and thereby improving the reliability and effectiveness of features; spatial attention mechanisms enable the network to focus on the feature importance in different positions of the image, enhancing the local accuracy and global consistency of the features.
    \item ResNet, as a classic deep residual network, demonstrates excellent performance in many visual tasks. Combining CBAM modules can further improve the model's accuracy and generalization performance.
    \item CBAM modules are compatible with the basic structure of ResNet, making them easy to integrate into existing network architectures.
    \item CBAM modules themselves have a relatively small computational cost and do not impose a significant additional burden on the network.
\end{enumerate}

\subsection{Model Lightening}

To maintain high performance while reducing the demand for computational resources, this section discusses three model lightening techniques: model pruning, knowledge distillation, and their combination.

\subsubsection{Model Pruning}

To perform model lightening through pruning, a pruning criterion must first be established. Commonly, the absolute value of weights serves as the criterion. The specific steps are as follows:

\begin{enumerate}
    \item Sort each element in the weight matrix $W$ according to its absolute value.
    \item Choose a threshold $t$ and set weights below the threshold to 0. In this manner, the pruned weight matrix $W'$ can be expressed as:

    \begin{equation}
    W'[i, j] = \begin{cases}
    W[i, j], & \text{if } |W[i, j]| > t \\
    0, & \text{otherwise}
    \end{cases}
    \end{equation}
    
    \item Replace the original weight matrix $W$ in the model with the pruned weight matrix $W'$ to obtain a lighter model.
\end{enumerate}

By employing the above steps, model lightening is achieved through pruning, reducing the model's computational load and storage requirements, and making deep learning models easier to deploy and run on resource-constrained devices. It is worth noting that the pruning process may lead to a decrease in model accuracy, necessitating a balance between the degree of pruning and model performance. Experiments can be conducted to adjust the threshold $t$ and pruning strategy to obtain a satisfactory lightweight model.

\subsubsection{Knowledge Distillation}

Knowledge distillation mainly consists of two components: a teacher model and a student model.

Assuming the teacher model is T and the student model is $S$, with input $x$, the output of the teacher model is $T(x)$ and the output of the student model is $S(x)$. The objective of knowledge distillation is to enable the student model to learn the knowledge of the teacher model by minimizing the output difference between the two models.

First, define a temperature parameter $T_d$ for smoothing the output probability distribution of the models. Then, calculate the output probability distributions of the teacher and student models using softmax with temperature:

\begin{equation}
    P_T(x) = Softmax(\frac{T(x)}{T_d})
\end{equation}

\begin{equation}
    P_S(x) = Softmax(\frac{S(x)}{T_d})
\end{equation}

Next, minimize the difference in probability distributions between the teacher and student models. The Kullback-Leibler (KL) divergence is used to measure this difference:

 \begin{equation}
     L_{KL} = KL(P_{T(x)} || P_{S(x)})
 \end{equation}
 
In addition, to ensure that the student model performs well on the labels, minimize the cross-entropy loss between the output probability distribution $P_{S(x)}$ and the true label $y$:

\begin{equation}
    L_{CE} = CrossEntropy(P_{S(x)}, y)
\end{equation}

Finally, linearly weight the two loss functions to obtain the final loss function:

\begin{equation}
    L = \alpha \times L_{KL} + (1-\alpha)\times L_{CE}
\end{equation}
 
Here, $\alpha$ is the weight coefficient used to balance the impact of KL divergence loss and cross-entropy loss. By minimizing the loss function $L$, the objective of knowledge distillation is achieved, and the student model learns the knowledge of the teacher model.

\subsubsection{Integrating Model Pruning and Knowledge Distillation}

As shown in the previous discussions, pruning and distillation are two different approaches to model lightweighting. In this section, the simultaneous use of these two techniques to further reduce the model size is discussed. The specific steps for integrating model pruning and knowledge distillation for model lightweighting are as follows:

First, model pruning is performed as described earlier, determining the pruning criteria and removing low-weight connections in the weight matrix to obtain the pruned weight matrix $W'$. The pruned model is used as the teacher model $T$. Next, a student model $S$ with a similar structure to the teacher model, but more lightweight, is initialized.

Then, following the steps of knowledge distillation, the teacher model $T$ and the student model $S$ are used for knowledge distillation training. During the training process, the student model $S$ learns the knowledge of the teacher model $T$ by minimizing the loss function $L$.

Finally, after knowledge distillation training, the student model $S$ will have a more lightweight structure and higher performance. At this point, the student model is applied to real tasks for high-precision, low-computational complexity object detection and classification. The distilled model is fine-tuned with the pruned model to better balance model size and accuracy.

The fine-tuning process typically uses a stochastic gradient descent optimizer and adjusts the weights of the distilled model according to the pruning ratio of the pruned model. Therefore, the goal of the fine-tuning process is to further reduce the model size while maintaining its accuracy. The following formula is used to update the weights of the fine-tuned model:

\begin{equation}
    w_{i+1}= w_{i}-\eta \nabla L(w_{i};D_{train})
\end{equation}

where $w_{i+1}$ is the weight after the $i+1$ iteration, $w_{i}$ is the weight after the $i$ iteration, $\eta$ is the learning rate, $\nabla L(w_{i};D_{train})$ is the gradient of the loss function, and $D_{train}$ is the training dataset.

Furthermore, to balance model size and accuracy, the following formula is used to adjust the weights of the distilled model:

\begin{equation}
    \hat{w}_{i+1}=\alpha \times w_{i+1}+(1-\alpha) \times \hat{w}_{i}
\end{equation}

where $\hat{w}_{i+1}$ is the weight of the distilled model after the $i+1$ iteration, $\hat{w}_{i}$ is the weight of the distilled model after the $i$ iteration, and $\alpha$ is the weight adjustment factor, typically ranging between 0.5 and 0.9.

By combining the above steps, a lightweight model is obtained, which has similar accuracy to the original model but fewer parameters and lower computational complexity.

\subsection{Label Smoothing}

To endow the model with better generalization capabilities, the method of label smoothing was employed. In deep learning, label smoothing is a regularization technique that can mitigate overfitting and enhance the model's generalization performance. Label smoothing is achieved by transforming the target label's hard distribution (one-hot encoding) into a softer distribution. Specifically, given the target class $y$ and smoothing factor $\alpha$, label smoothing is calculated as follows:

\begin{equation}
    \tilde{y_{i}} = \begin{cases}
    (1-\alpha)+\frac{\alpha}{K}, & \text{if i=y} \\
    \frac{\alpha}{K}, & \text{otherwise}
    \end{cases}
\end{equation}

Where $i$ is the class index, $K$ is the total number of classes, and $\tilde{y}_i$ is the smoothed target distribution.

\section{Results And Analyses}
\subsection{Datasets}

The apple maturity and damage dataset used in this study was collected in the autumn of 2022. A Canon 5D camera was employed during the data collection process to ensure image quality. A total of 3000 apple images were collected, including samples with different maturity levels and degrees of damage.

\subsection{Experiment}

\label{sec:Expeiment}

\subsubsection{Evaluation Metrics}

Since the task of this study is apple classification and defect detection, the evaluation metrics include classification metrics and object detection metrics. Evaluation metrics are used to measure the performance and predictive ability of the model. The following evaluation metrics were chosen:Recall, Precision, Accuracy, F1-score, FPS and mAP.
mAP is a metric that measures the performance of an object detection model. It calculates the average precision at different recall rates, reflecting the model's performance across the entire dataset. 


\subsubsection{Experimental Settings}

In the hardware setup for this experiment, the GPU used is an NVIDIA GeForce RTX 3080. Using a graphics card with a large memory capacity can better handle large datasets and train complex models. The CPU is an Intel Core i7 series processor. A processor with a higher number of cores and threads delivers better performance. 32 GB of memory is used to support the processing of large datasets and models. Storage consists of a 1 TB SSD to store large datasets and experimental results.

In the software setup for this experiment, the operating system is Ubuntu 18.04, as many deep learning libraries and tools are easier to install and configure in a Linux environment. The programming language is Python 3.6, and the deep learning framework is TensorFlow 2.x. These frameworks support automatic differentiation, GPU acceleration, and the implementation of various deep learning models. Auxiliary libraries and tools include NumPy, SciPy, Pandas, Matplotlib, and other Python libraries for data processing, scientific computing, and visualization. OpenCV is used for image processing and computer vision tasks. CUDA and cuDNN are used for GPU computation acceleration.

\subsection{Apple Grading Results}

In this section, to validate the practical application performance of the proposed model on the apple grading problem involving ripe, unripe, and defective apples, experiments were conducted and the results were compared with those of other models using evaluation metrics such as precision, recall, accuracy, and FPS. In the benchmark system, the GoogLeNet, whose main parameters include network depth (22 layers), loss function (cross-entropy), optimizer (stochastic gradient descent), and learning rate (initial value set to 0.01, decaying by 10\% every 5 epochs). The weight decay coefficient was set to 0.0002 and the momentum to 0.9. For the DenseNet model, the main parameters include network depth (121 layers), loss function (cross-entropy), optimizer (stochastic gradient descent), learning rate (initial value set to 0.1, decaying by 10\% every 30 epochs). The weight decay coefficient was set to 0.0001 and the momentum to 0.9. The main parameters of the ZFNet include network depth (8 layers), loss function (cross-entropy), optimizer (stochastic gradient descent), and learning rate (initial value set to 0.01, decaying by 10\% every 5 epochs). The weight decay coefficient was set to 0.0005 and the momentum to 0.9. For the ResNet, the main parameters include network depth (50 layers), loss function (cross-entropy), optimizer (stochastic gradient descent), learning rate (initial value set to 0.1, decaying by 10\% every 30 epochs). The weight decay coefficient was set to 0.0001 and the momentum to 0.9. The main parameters of the AlexNet model include network depth (8 layers), loss function (cross-entropy), optimizer (stochastic gradient descent), and learning rate (initial value set to 0.01, decaying by 10\% every 10 epochs). The weight decay coefficient was set to 0.0005 and the momentum to 0.9. For the VGG model, the main parameters include network depth (19 layers), loss function (cross-entropy), optimizer (stochastic gradient descent), learning rate (initial value set to 0.01, decaying by 10\% every 20 epochs). The weight decay coefficient was set to 0.0005 and the momentum to 0.9. The results are shown in Table \ref{tab: apple grading results}.

\begin{table}[H]
    \centering
	\caption{Comparison of different models for the apple grading Problem. 
    \label{tab: apple grading results}}
	\begin{tabular}{lcccc}
		\toprule
		\textbf{Model}   & \textbf{Precision}  & \textbf{Recall}& \textbf{Accuracy} &\textbf{FPS}\\
		\midrule
         GoogleNet       & 87.5                & 86.6            & 87.2              & 48.2 \\
         DenseNet        & 89.4                & 87.3            & 88.7              & 50.3 \\
         ZFNet           & 90.3                & 88.6            & 89.4              & 53.9 \\
         ResNet          & 92.6                & 91.7            & 92.2              & 52.7 \\
         AlexNet         & 84.7                & 83.5            & 84.1              & 46.9 \\
         VGG             & 87.4                & 86.2            & 86.8              & 44.1 \\
         Proposed model  & 95.6                & 93.8            & 95.0              & 56.5 \\
		\bottomrule
	\end{tabular}
\end{table}

According to the data in the table, it can be concluded that the proposed model outperforms other typical deep learning models, such as GoogleNet, DenseNet, ZFNet, ResNet, AlexNet, and VGG, in the apple classification problem. The proposed model achieves the best performance in evaluation metrics such as precision, recall, accuracy, and FPS (frames per second).

The proposed model achieves a precision of 95.6\%, which is significantly higher than that of other models, with ResNet reaching the second-highest value of 92.6\%. In terms of recall, the proposed model achieves the highest level of 93.8\%, with ResNet following closely at 91.7\%. This indicates that the proposed model has a stronger ability to cover true positive cases. In terms of accuracy, the proposed model achieves 95.0\%, outperforming other models, with ResNet being the second-highest at 92.2\%. This further demonstrates the excellent performance of the proposed model in the ratio of correctly classified samples to the total number of samples. Finally, in terms of FPS, the proposed model achieves the highest value of 56.5, with ZFNet being the second-highest at 53.9. This implies that the proposed model has a faster processing speed in terms of real-time performance.

\subsection{Apple Defect Detection Results}

In this section, to verify the practical application performance of the proposed model in apple defect detection, experiments were conducted, and the model was compared with other models using precision, recall, and MAP evaluation metrics.In the benchmark system, for the SSD, key parameters include network depth (16 layers), loss functions (Smooth L1 Loss and Cross Entropy), optimizer (Stochastic Gradient Descent), learning rate (initially set to 0.001 and decays by 10\% every 10 epochs), weight decay coefficient (0.0005), and momentum (0.9). In the YOLOv3, key parameters comprise network depth (106 layers), loss functions (Mean Square Error and Binary Cross Entropy), optimizer (Adam), learning rate (initially set to 0.001 and decays by 10\% every 20 epochs), and weight decay coefficient (0.0005). Regarding the YOLOv4, principal parameters include network depth (110 layers), loss functions (Mean Square Error and Binary Cross Entropy), optimizer (Adam), learning rate (initially set to 0.001 and decays by 10\% every 20 epochs), and weight decay coefficient (0.0005). In the YOLOv5, key parameters encompass network depth (67 layers), loss functions (Mean Square Error and Binary Cross Entropy), optimizer (Adam), learning rate (initially set to 0.01 and decays by 10\% every 50 epochs), and weight decay coefficient (0.0005). For the FSSD, main parameters include network depth (ResNet50), loss functions (Smooth L1 Loss and Cross Entropy), optimizer (Stochastic Gradient Descent), learning rate (initially set to 0.001 and decays by 10\% every 10 epochs), weight decay coefficient (0.0005), and momentum (0.9). In the Faster R-CNN, primary parameters consist of network depth (ResNet50), loss functions (Smooth L1 Loss and Cross Entropy), optimizer (Stochastic Gradient Descent), learning rate (initially set to 0.001 and decays by 10\% every 10 epochs), weight decay coefficient (0.0001), and momentum (0.9). For the EfficientDet, main parameters include network depth (EfficientNet-B7), loss functions (Mean Square Error and Binary Cross Entropy), optimizer (Stochastic Gradient Descent), learning rate (initially set to 0.1 and decays by 10\% every 2 epochs), and weight decay coefficient (0.0001), and momentum (0.9). The specific data is shown in the following table:

\begin{table}[H]
    \centering
	\caption{Comparison of object detection models for the apple defect detection. 
    \label{tab: apple defect detection}}
	\begin{tabular}{lccc}
		\toprule
		\textbf{Model}   & \textbf{Precision}  & \textbf{Recall} & \textbf{mAP} \\
		\midrule
         Faster R-CNN    & 84.5                & 82.3            & 83.2 \\
         SSD             & 87.8                & 86.1            & 87.2 \\
         YOLO v3         & 85.4                & 84.2            & 84.7 \\
         YOLO v4         & 89.3                & 87.9            & 88.6 \\
         YOLO v5         & 93.2                & 91.8            & 92.5 \\
         EfficientDet    & 94.3                & 92.9            & 93.7 \\
         FSSD            & 83.1                & 81.2            & 82.6 \\
         Proposed model  & 95.3                & 93.7            & 94.5\\
		\bottomrule
	\end{tabular}
\end{table}

It can be observed from the table that the accuracy performance of different object detection models varies under different input sizes. At an input size of 300*300, YOLO v5 has the highest accuracy, achieving 93.2\% Precision, 91.8\% Recall, and 92.5\% mAP. In contrast, Faster R-CNN and SSD perform relatively average, with only 83.2\% and 87.2\% mAP, respectively. This also indicates that the performance of object detection models is not only related to the model's design but also greatly affected by the input image size. In different application scenarios, a suitable model and input size need to be selected according to specific requirements.

\subsection{Impact of Different Pruning Strategies on Model Accuracy}

In this section, the impact of various pruning strategies on model accuracy is discussed. Ablation experiments were conducted for four different schemes, which are as follows:

\begin{enumerate}
    \item No pruning: The unpruned model serves as the baseline, with its accuracy, recall, and computational overhead evaluated on the test set.
    \item Random pruning: The simplest random pruning approach is employed, in which certain neuron or channel weights are randomly set to zero to reduce the number of model parameters.
    \item Sparse pruning: Pruning methods based on sparse matrices, such as L1 or L2 regularization, are used to set certain weights to values close to zero, further reducing the number of model parameters.
    \item Distillation pruning: A combination of model pruning and knowledge distillation is used for the ablation experiments. Specifically, a pruning method is first employed to reduce the number of model parameters, and then a pre-trained model is used as the "teacher model" to train the pruned model with its output results as "soft labels."
\end{enumerate}
 
The experimental results are shown in the following table:

\begin{table}[H]
    \centering
    \caption{Model pruning ablation experiment results.}
    \label{tab:model pruning ablation}
    \begin{tabular}{lccc}
        \toprule 
        \textbf{Model}        & \textbf{Precision} & \textbf{Recall} & \textbf{mAP} \\
        \midrule
        No pruning            & 93.1               & 92.4            & 91.3 \\
        Random pruning        & 94.3               & 93.2            & 93.8 \\
        Sparse pruning        & 92.9               & 90.3            & 91.5 \\
        Distillation pruning  & 95.3               & 93.7            & 94.5 \\
        \bottomrule
    \end{tabular}
\end{table}

From a mathematical perspective, random pruning reduces the size of the model by randomly deleting some parameters in the network, but it may also remove some key parameters, leading to a decline in model performance. Sparse pruning is another pruning method that reduces the size of the model by setting smaller weights to zero. However, if useful weights are removed, the performance of the model may decline. In contrast, distillation pruning is a pruning method based on knowledge distillation. It uses the knowledge of the teacher model to guide the learning of the student model, resulting in a smaller, faster, and more accurate model. Therefore, compared with other pruning methods, it can be seen that distillation pruning can significantly reduce the size of the model while maintaining high performance. Distillation pruning can improve the computational efficiency of the model by eliminating unnecessary parameters, while maintaining accuracy comparable to the unpruned model. This is because distillation pruning is not simply trimming network parameters, but more effectively training a smaller model through knowledge distillation, making better use of known information.

For random pruning, the experimental results show that its precision, recall, and mAP are 94.3\%, 93.2\%, and 93.8\%, respectively, all of which are better than the unpruned model. This suggests that the random pruning method can slightly improve the performance of the model while reducing the complexity of the model. This may be because the pruning process eliminates some redundant and unimportant parameters, making the model more focused on important features, thereby improving the generalization ability of the model. Next, for sparse pruning, although its precision of 92.9\% is slightly lower than that of the unpruned model, overall, its performance has not been significantly affected, which proves the effectiveness of sparse pruning. However, its recall and mAP are 90.3\% and 91.5\% respectively, lower than the unpruned model. This may be because sparse pruning may also cut off some parameters that contribute to the performance of the model in the process of reducing parameters. Finally, for distillation pruning, its precision, recall, and mAP are 95.3\%, 93.7\%, and 94.5\% respectively, all of which are significantly better than the unpruned model. This shows that the distillation pruning method can effectively reduce the complexity of the model while maintaining model performance. This is due to the characteristics of distillation pruning, which not only considers the importance of parameters but also considers the redundancy of parameters, so it is more focused on maintaining those parameters that are most important to model performance during the pruning process.

\subsection{Impact of Different Attention Modules on Model Accuracy}

In this section, the impact of various attention modules on model accuracy is discussed. The experimental results are shown in the following table:

\begin{table}[H]
    \centering
    \caption{Detection accuracy of different attention modules.}
    \label{detection accuracy}
    \begin{tabular}{lccc}
        \toprule 
        \textbf{Model}        & \textbf{Precision} & \textbf{Recall} & \textbf{mAP} \\
        \midrule
        CBAM                  & 95.3               & 93.7            & 94.5 \\
        SAM                   & 93.2               & 92.1            & 92.8 \\
        BAM                   & 94.8               & 93.3            & 93.9 \\
        \bottomrule
    \end{tabular}
\end{table}

According to the table data, it is evident that the CBAM module achieves the best results in terms of precision, recall, and mean average precision (mAP) for object detection compared to the SAM and BAM modules. This may be due to the CBAM module's design, which combines channel attention (CA) and spatial attention (SA) mechanisms, effectively utilizing the advantages of both.

\subsection{Visualization Analysis}

\begin{figure}[H]
    \includegraphics[width=14cm]{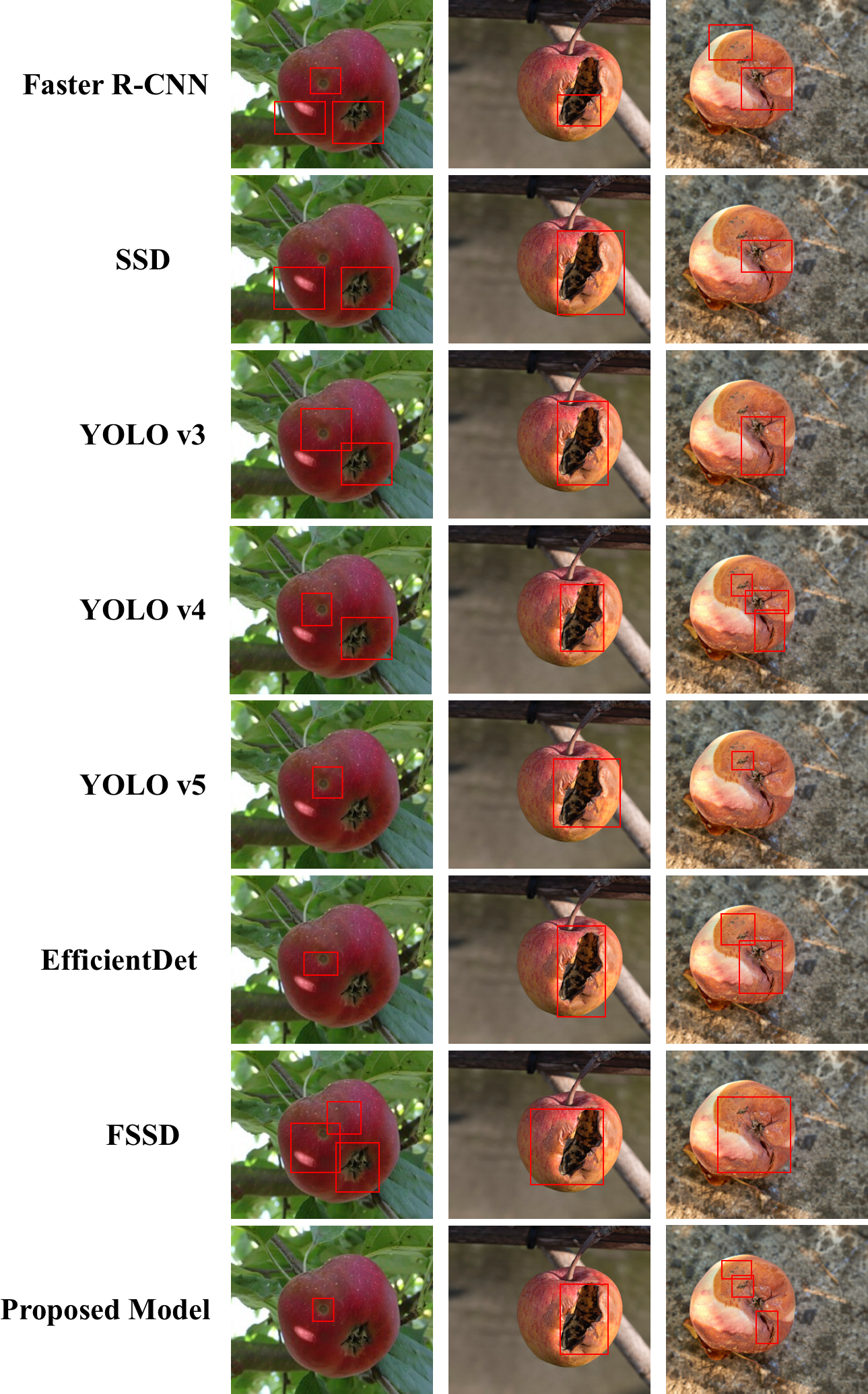}
    \caption{Images of different models in the apple defect detection task.}
    \label{fig: detection}
\end{figure}

In order to further explore the influencing factors of apple defect detection and enhance the generalization ability of the model, we conducted visualization operations, with specific contents as follows in Figure \ref{fig: detection}.

Based on the experimental results and generated images of different models in the apple defect detection task in the table above, the performance of each model can be described and analyzed. Influencing factors include lighting, shooting angle, defect size, and defect shape.

Faster R-CNN performs well in detecting apple defects under good lighting conditions, but its detection effect is affected under insufficient or uneven lighting conditions. Meanwhile, changes in defect shape and shooting angle also lead to a decrease in detection performance. SSD performs better under poor lighting conditions compared to Faster R-CNN. However, its detection effect is still affected when facing different defect shapes and shooting angles. In addition, SSD has limited detection ability for smaller defect spots. Under poor lighting conditions, varied defect shapes, and changes in shooting angles, the YOLO v3 model performs better than Faster R-CNN and SSD. However, its detection accuracy may decrease when the spot size is small. Compared to YOLO v3, YOLO v4 performs better under poor lighting conditions, occlusions, and changes in shooting angles. At the same time, the model has a strong detection ability for smaller defect spots. Under various lighting conditions, the YOLO v5 model can maintain high detection accuracy. At the same time, the impact of shooting angle is relatively small, and it has stronger robustness for spot size. In the face of insufficient light, occlusion, and changes in shooting angle, the EfficientDet model has relatively stable detection performance. In addition, the model has higher detection accuracy for smaller defect spots. Although FSSD has certain detection capabilities under good lighting conditions, its detection effect is poor under poor lighting, occlusion, and changes in shooting angles. At the same time, FSSD has weaker detection ability for smaller defect spots. Under various lighting conditions, our model performs excellently and has high detection accuracy. At the same time, the impact of different defect shapes and shooting angles is relatively small, and it has strong robustness for spot size.

In summary, by comparing the performance of each model in terms of lighting, defect shape, shooting angle, and defect size, our model has obvious advantages in the apple defect detection task. Under different conditions, our model can still maintain high detection accuracy and has a strong detection ability for smaller defect spots. These advantages are attributed to the lightweight convolutional neural network and generative adversarial network we adopted, which effectively solve the problems of sample imbalance and high computational resource requirements, thereby improving the real-time performance and recognition ability of the model in practical application scenarios. These advantages provide more intelligent solutions for the development of the agricultural industry.

\section{Discussion and Conclusion}

In this research, we propose a method based on a lightweight Convolutional Neural Network (CNN) and Generative Adversarial Network (GAN) for detecting apple ripeness and damage levels. This method achieves rapid, accurate, and automated detection of apples by employing a lightweight model design, utilizing GANs for data augmentation, and applying object detection networks for damage location annotations. Moreover, the development of attention mechanisms and the benefits of lightweight model techniques have been discussed in detail.

In terms of model design, by optimizing the depth and width of the model and using advanced model compression techniques, we propose a lightweight CNN model. This successfully reduces the model's parameter and computational demands, thereby enhancing its real-time performance in practical application scenarios. Additionally, we have introduced attention mechanisms that dynamically adjust the importance of different feature layers, thereby improving the model's performance in object detection tasks. To address the issues of imbalanced and insufficient samples in apple ripeness and damage detection tasks, we employ GANs to generate realistic apple images, thereby expanding the training dataset and enhancing the model's recognition capabilities for apples of varying ripeness and damage levels. Furthermore, by applying object detection networks to annotate damage locations on damaged apples, we improve the accuracy of damage level detection, providing a more accurate basis for decision-making.

Experiments conducted on an actual dataset demonstrate the advantages of our proposed lightweight CNN and GAN in improving apple classification and detection accuracy and reducing computational resource demands. In apple grading detection, the model achieves a precision of 95.6, a recall of 93.8, an accuracy of 95.0, and an FPS of 56.5. In apple defect detection, the model achieves a precision of 95.3, a recall of 93.7, and an mAP of 94.5. In both tasks, our proposed method outperforms other mainstream models. The experimental results indicate that our proposed method achieves satisfactory performance in apple ripeness and damage level detection tasks, demonstrating significant practical value.

The importance of this study, in terms of practical implications, is mainly manifested in the following aspects: Firstly, the model proposed in this research achieves precise automated detection of apple ripeness and damage levels. This holds significant practical significance for agricultural production. Traditional apple ripeness and damage level detection mainly relies on manual inspections, which are inefficient and prone to errors. Our proposed model can perform these tasks automatically, greatly improving detection efficiency and reducing human errors. This holds essential practical value in enhancing agricultural production efficiency and ensuring the quality of agricultural products. Secondly, by introducing GANs for data augmentation, we address the problem of insufficient and imbalanced training data. This implies that even in a data-scarce environment, our model can maintain high detection accuracy. This is of great importance in practical applications as, in many instances, acquiring sufficient training data is challenging. Furthermore, the model proposed in our research requires minimal computational resources and performs efficiently in real-time, making it suitable for practical application scenarios. This is especially important in agricultural applications where rapid detection is required in an environment with limited computational resources. Lastly, the structure and training strategies of our model are somewhat generic, which can be transferred to other crops or other object detection tasks. This significantly enhances the practical value of this research, making its application range broader within the agricultural industry.

In conclusion, this study not only theoretically promotes the development of lightweight CNNs and GANs, but also provides an effective method for detecting apple ripeness and damage levels in practical applications. It holds substantial practical significance in improving agricultural production efficiency, ensuring the quality of agricultural products, and addressing the problem of data scarcity. Future research will build on this foundation, further exploring a variety of application scenarios, providing more intelligent solutions for the development of the agricultural industry.

\bibliographystyle{unsrt}  
\bibliography{references}  





\end{document}